\pdfoutput=1
\documentclass[11pt]{article}

\usepackage[final]{acl}
\usepackage{comment}
\usepackage[framemethod=TikZ]{mdframed}
\usepackage{times}
\usepackage{latexsym}
\usepackage{subcaption}
\usepackage{caption}
\usepackage{float}

\usepackage[T1]{fontenc}
\usepackage[utf8]{inputenc}

\usepackage{microtype}

\usepackage{inconsolata}

\usepackage{graphicx}
\usepackage{multirow}
\usepackage{booktabs}
\usepackage{adjustbox}
\DeclareUnicodeCharacter{2714}{\checkmark}
\usepackage{xspace}
\usepackage{pifont}
\usepackage{xcolor}
\usepackage{tikz}

\definecolor{customyellow}{HTML}{FFBF00}
\newcommand*\circled[1]{%
  \tikz[baseline=(char.base)]{
    \node[shape=circle,draw=black,fill=customyellow,,inner sep=1pt,font=\scriptsize] (char) {\strut #1};}}

\newcommand{\cmark}{\textcolor{green}{\ding{51}}}
\newcommand{\xmark}{\textcolor{red}{\ding{55}}}
\newcommand{\toolname}{\textsc{CowPilot}} 
\newcommand{\gpt}{\textsc{GPT}\xspace}
\newcommand{\llama}{\textsc{LLaMa}\xspace}

\title{
    \adjustbox{valign=c}{\includegraphics[height=2em]{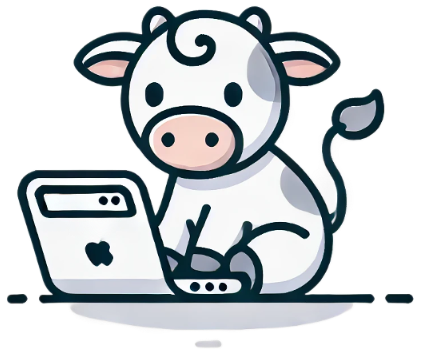}} 
    \toolname{}: A Framework for Autonomous and Human-Agent\\ Collaborative Web Navigation
}

\author{
  \textbf{Faria Huq \quad Zora Zhiruo Wang \quad Frank F. Xu \quad Tianyue Ou \quad Shuyan Zhou}\\
  \textbf{Jeffrey P. Bigham\textsuperscript{\textdagger} \quad Graham Neubig\textsuperscript{\textdagger}} \\
  School of Computer Science, Carnegie Mellon University \\
  \texttt{\{fhuq, jbigham, gneubig\}@cs.cmu.edu} \\
  \textsuperscript{\textdagger}Equal Supervision \\
  \texttt{Code:} \url{https://github.com/oaishi/CowPilot}
}

\begin{document}
\maketitle

\begin{abstract}
While much work on web agents emphasizes the promise of autonomously performing tasks on behalf of users, in reality, agents often fall short on complex tasks in real-world contexts and modeling user preference. This presents an opportunity for humans to collaborate with the agent and leverage the agent’s capabilities effectively. We propose \toolname{}, a framework supporting autonomous as well as human-agent \underline{co}llaborative \underline{w}eb navigation, and evaluation across task success and task efficiency.  \toolname{} reduces the number of steps humans need to perform by allowing agents to propose next steps, while users are able to pause, reject, or take alternative actions. During execution, users can interleave their actions with the agent’s by overriding suggestions or resuming agent control when needed. %\toolname{} eases human effort by starting with agents proposing next steps, meanwhile allowing humans to execute, pause, or reject agent-proposed steps and take alternative actions, allowing for interleaved action-taking between agent and human participants. 
We conducted case studies on five common websites and found that the human-agent collaborative mode achieves the highest success rate of $95\%$ while requiring humans to perform only $15.2\%$ of the total steps. Even with human interventions during task execution, the agent successfully drives up to half of task success on its own. \toolname{} can serve as a useful tool for data collection and agent evaluation across websites, which we believe will enable research in how users and agents can work together. Video demonstrations are available at \url{https://oaishi.github.io/cowpilot.html}

\end{abstract}

\section{Introduction}

Large language models (LLMs) have shown remarkable capabilities in automating digital tasks involving complex decision-making and information retrieval processes \citep{zhou2023webarena, deng2024mind2web}. Despite these advancements, measuring the effectiveness of human-agent collaboration remains a challenging endeavor.

Previous studies have primarily focused on measuring task success and efficiency without adequately addressing the nuanced interactions between users and agents. In this follow-up work, we aim to address the following research questions:

\begin{enumerate}
    \item How can we improve the evaluation of human-agent collaboration in web navigation tasks?
    \item What factors contribute to successful or unsuccessful collaborations?
    \item How do users' interventions influence the overall effectiveness of the system?
\end{enumerate}

% Agents supported by large language models (LLMs) have become increasingly capable of automating digital tasks such as web navigation \citep{zhou2023webarena, deng2024mind2web}. While existing frameworks for web agents mostly focus on solo, autonomous agents \cite{webolympus, openwebagent, drouin2024workarena}, we argue that, for many practical tasks, users interact with the LLM agent for varied purposes such as supervision and collaboration, i.e., the \textit{copilot} mode. 
While existing frameworks \citep{lu2024weblinx,drouin2024workarena,wang2024openhands,webolympus} mainly support users communicating with agents via natural language (NL) feedback, or recording actions of human users alone \citep{pan2024webcanvas}, they do not support dynamic human-agent collaboration within a task session, where humans and LLM agents take actions alternately to solve tasks. We ask: \textit{How can we enable human-agent collaborative task-solving?} and further, \textit{how do agents perform under such collaborative settings?}

To facilitate studying these questions, we introduce \toolname{} (\S\ref{sec:2:method}), a lightweight framework that can be seamlessly integrated into user web activities as a Chrome extension. 
\toolname{} starts with the LLM agent \textit{propos}ing actions for the next step, meanwhile allowing human to \textit{pause} or reject the agent-suggested action and take alternative ones to drive the process; human can also choose to \textit{resume} the agent-driven process at any time to ease the effort (\S\ref{sec:2.1:system}). To systematically evaluate this collaborative process, we propose several metrics for task accuracy, user experience, and efficiency aspects (\S\ref{sec:2.2:eval-metrics}).

\begin{figure*}
    \centering
    \includegraphics[width=\linewidth]{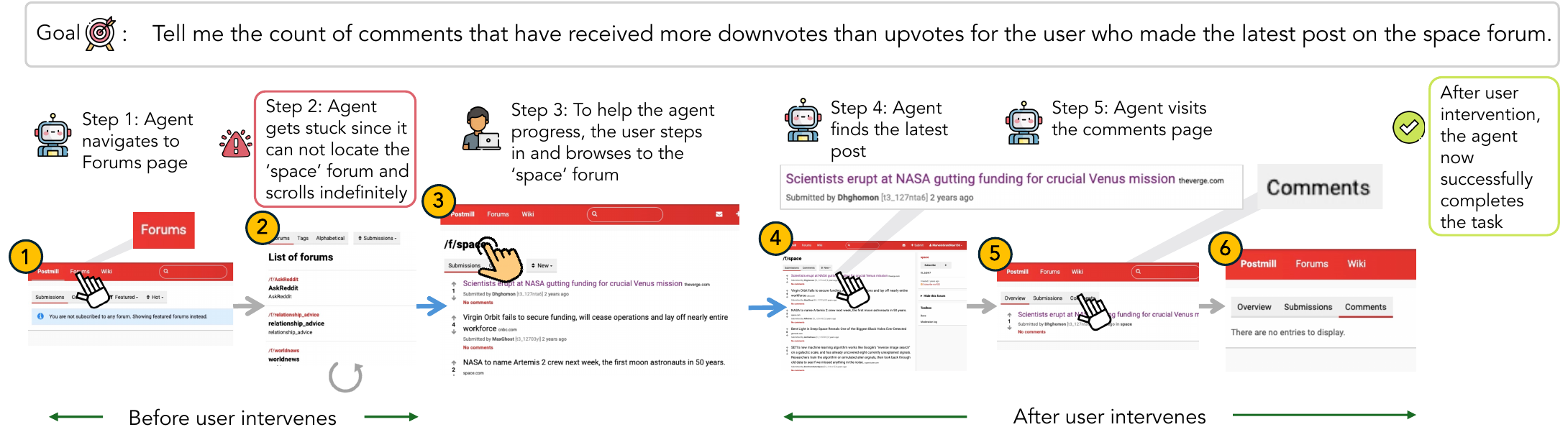}
    \caption{A step-by-step illustration of how human intervention enables the agent to overcome a failure point during task execution. The figure uses gray edges to represent the agent's autonomous actions and blue edges to indicate human intervention. The process begins with the agent attempting the task independently (Step  \protect\circled{1}) and navigating to the interface to list available forums (Step \protect\circled{2}). At this stage, the agent gets stuck, unable to locate the desired `space' forum. A human intervenes (Step \protect\circled{3}), guiding the agent to the correct forum. The user then resumes the agent's operation (Step \protect\circled{4}), allowing it to retrieve the required post and complete the task by navigating to the comments section (Step \protect\circled{5})."
    }
\end{figure*}

Beyond agent web automation, \toolname{} enables a wide range of use cases (\S\ref{sec:3:usecase}), including web automation (\S\ref{subsec:task1}), data collection for agent trajectories and user feedback (\S\ref{subsec:task2}), as well as evaluations for single or multiple agents (\S\ref{subsec:task3}).

We conduct studies on five common websites across shopping, social, and technical domains (\S \ref{sec:results}). We show \toolname{} in collaborative mode achieves higher success rates over autonomous agents by $47\%$, and even human-only settings by $6\%$. Moreover, the LLM agent takes $84.8\%$ of the steps and can drive up to half of the task success, greatly easing human efforts. These results suggest the potential for accuracy and efficiency improvement with \toolname{}.

Overall, \toolname{} showcases the potential of human-agent collaborative web navigation, and serves as a useful tool for future web automation, data collection, and evaluation research.
\section{\toolname{}}
\label{sec:2:method}
In this section, we introduce the \toolname{} framework (\S\ref{sec:2.1:system}) and evaluation metrics for task accuracy and collaboration quality (\S\ref{sec:2.2:eval-metrics}).

\subsection{The \toolname{} System}
\label{sec:2.1:system}
Given an objective $o$ stated in natural language (NL) (e.g., \textit{book a flight from New York to Pittsburgh}) for the web environment, we define two agents: one agent instantiated with an LLM policy $\pi_{\mathcal{L}}$, and one human agent $\pi_{\mathcal{H}}$. At each task step $t$, based on the observation $o_t$ from the environment state $s_t$, either the LLM agent or human agent generates an action $a_t$, formalized as $a_t = \pi(t, o_t, a_{0:t-1})$. Executing $a_t$ on the environment results in a new state $s_{t+1}$ that gives observation $o_{t+1}$ that drives the next step. The two agents collectively generate a sequence of actions $a_{0:n}$ over $n$ steps, until it reaches a task termination condition, e.g., output \texttt{STOP} or a maximum number of steps.
By default, the LLM agent starts generating actions $a_0^{\mathcal{L}}$ from $t=0$ unless intervened by the human agent $\pi_{\mathcal{H}}$. \autoref{tab:action-space} shows the action space for LLM and human agents.

\begin{table*}[ht]
\centering
\resizebox{0.93\textwidth}{!}{%
\begin{tabular}{lll}
\toprule
\multicolumn{1}{l}{\bf Action} & \multicolumn{1}{c}{\bf Raw Human Action} & \multicolumn{1}{l}{\bf Description} \\
\midrule
\texttt{click(elem)} & \texttt{click} & Click on a webpage element using the mouse. \\
\texttt{hover(elem)} & \texttt{mouseover} & Hover the mouse over an element without clicking it. \\
\texttt{type(elem, text)} & \texttt{input} & Enter text into a text area or text box. \\
\texttt{scroll(dir)} & \texttt{wheel} & Scroll the webpage up/down/left/right. \\
\midrule
\texttt{goto(url)} & \texttt{Tabs.onUpdated} & Navigate to a specific URL. \\
\texttt{goto(tab)} & {-} & Navigate to a specific tab. \\
\midrule
\texttt{finishwithanswer(text)} & {-} & \begin{tabular}[c]{@{}l@{}}For information retrieval task, terminate the task with \\ retrieved textual information.\end{tabular} \\
\texttt{finish()} & {-} & Mark the task as completed. \\
\texttt{failure()} & {-} & Mark the task as failed and uncompleted.\\
\bottomrule
\end{tabular}
}
\caption{Action space of agents in \toolname{}. LLM agent supports all actions in the \textit{Action} column. Human actions are captured by entries in the \textit{Raw Human Action} column and transformed into \textit{Action}s.}
\label{tab:action-space}
\end{table*}

\begin{figure*}
    \centering
    \includegraphics[width = 0.8\textwidth]{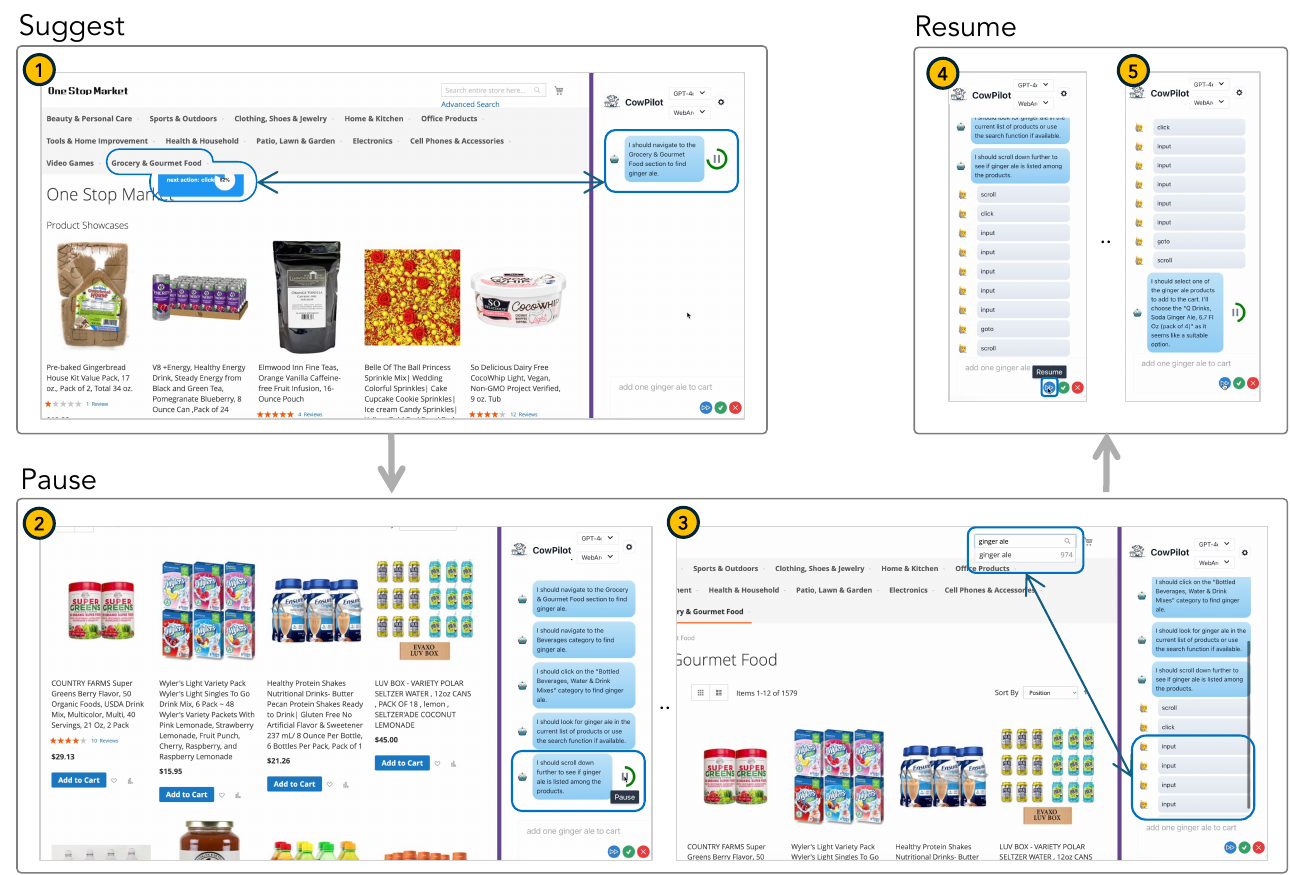}
    \caption{Example of \toolname{}'s core interaction modules during task execution. At step \protect\circled{1}, the LLM agent generates a \textbf{\textit{suggestion}}, highlighting the textual description and the UI element where the action will be performed. At step \protect\circled{2}, the user identifies an erroneous action, chooses to \textbf{\textit{pause}} the LLM agent, and proceeds to perform corrective actions manually (step \protect\circled{3}, e.g., typing in the textfield, highlighted in blue). At step \protect\circled{4}, the user chooses to \textit{\textbf{resume}} the LLM agent, allowing it to continue generating actions. The agent resumes successfully and proceeds to execute subsequent steps autonomously (step \protect\circled{5}).}
    \label{fig:breakdown}
\end{figure*}

Actions taken by the human agent $\pi_{\mathcal{H}}$ are critical for optimizing \toolname{}'s decision-making pipeline. When the human agent intervenes, they provide contextual feedback by identifying and correcting prior mistakes made by the LLM agent. This redirection helps the agent recover from a suboptimal path and proceed with a more viable course of action.
At the same time, by integrating human actions into its action history, \toolname{} ensures that the LLM agent is aware of human corrections since its last decision, preventing redundant actions and enabling efficient task progression. To ensure effective integration of these human actions, \toolname{} incorporates the following core modules:

\noindent \textbf{Suggest-then-Execute under Human Supervision} \quad 
At any time step, the human agent $\pi_{\mathcal{H}}$ can decide to take over by generating action $a_t^{\mathcal{H}}$.
More concretely, the LLM agent $\pi_{\mathcal{L}}$ generates an action $a_t^{\mathcal{L}}$ and presents it as a \textit{suggestion} 
for the tentative next step to the user (\autoref{fig:breakdown}, \protect\circled{1}), which includes a visual indicator highlighting the target element for the proposed action, accompanied by a textual explanation of the agent's reasoning. This tentative step is presented to the human agent for at most five seconds, and is automatically executed if the human agent does not oppose. Otherwise, the human agent can choose to \textit{reject} or \textit{pause} the action (\autoref{fig:breakdown}, \protect\circled{3}) and take over. They can also transfer the action back to the LLM-based agent by hitting the \textit{resume} button (\autoref{fig:breakdown}, \protect\circled{4}). This take-over-then-back process can be conducted unlimited times per task-solving session. This mechanism balances operational efficiency with user oversight, allowing users to intercept potential errors without the burden of manually approving every step.

\noindent \textbf{Pause LLM Agent: Extract Human Actions} \quad  Whenever the human agent $\pi_{\mathcal{H}}$ rejects the LLM-proposed action, our \toolname{} system starts tracking human activity on the websites, particularly what webpages and UI elements they interact with. To capture this micro-level metadata, we utilize HTML event listener\footnote{\url{https://developer.mozilla.org/en-US/docs/Web/API/EventTarget/addEventListener}}, that are attached to the interactive elements (\textit{e.g.}, text field, buttons, dropdown menu) in the current webpage and triggered each time the elements are accessed by the user. To efficiently transfer user log data to the Chrome extension popup, we employ Chrome Message passing\footnote{\url{https://developer.chrome.com/docs/extensions/develop/concepts/messaging}} with a custom storage mechanism that is accessible by both the React AppState and \texttt{chrome.localStorage}\footnote{\url{https://developer.chrome.com/docs/extensions/reference/api/storage}}.

Note that, actions captured by the HTML event listener can include noisy actions irrelevant to the task \cite{Siqueira2018LeveragingAO, Cheng2015ProcessMO}. Hence, we transform the listener-captured actions to the LLM agent actions space, from \textit{Raw Human Action} to \textit{Action} column in \autoref{tab:action-space}. \textit{Action} also includes a textual description that helps the agent interpret user inputs effectively. To facilitate this transformation, we use an off-the-shelf LLM (\texttt{gpt-4o-2024-08-06}) and provide the raw human actions as input. The model outputs the transformed and cleaned version of the actions. (Find the prompt for this transformation in \textsection \ref{subsec:trans}.)

\noindent \textbf{Resume LLM Agent: Predict next Action using Human Input} \quad 
If the human agent chooses to \textit{resume} the LLM agent at any given step, \toolname{} stops tracking human actions and restarts LLM agent generation (\autoref{fig:breakdown}). Note that the LLM agent has access to all previous actions generated by itself and the human agent.

\subsection{Evaluation Metrics}
\label{sec:2.2:eval-metrics}
To evaluate the agent performance in \toolname{}, we report general agent task success. In addition, to better quantify human-agent collaboration, we introduce five evaluation metrics to measure various aspects throughout task execution.

\paragraph{General Task Success}
To measure generic task success, we measure \textit{end-to-end task accuracy}, which measures if the task objective is achieved after the agent task-solving process. At the end of the task, the agent self-marks its success or failure by generating a \texttt{finish} or \texttt{failure} action highlighted in \autoref{tab:action-space}. Optionally, the user can overwrite the success measure if they disagree with the agent's self-evaluation.

\paragraph{Human-Agent Collaboration}
To measure how human and agent interacted with each other throughout the task execution, we first measure the engagement of both parties, by: (1) \textit{Agent step count}: how many steps are taken by the agent per task; (2) \textit{Human step count}: how many steps are taken by the human per task; (3) \textit{Total step count}: the sum of steps taken by human and agent. 

Meanwhile, we measure agent capabilities via
(4) \textit{Human intervention count}: how many times does the user pause the agent to take actions themselves. Note that a single intervention may involve multiple steps performed by the human, as the intervention continues until the agent resumes. A higher value potentially suggests that the agent made frequent errors and users had to step in to resolve the mistakes;
and (5) \textit{Agent-driven completion accuracy}: how many tasks are completed by the agent, i.e., the terminating step was taken by the agent. A higher value indicates the agent's ability to recover and complete tasks autonomously after human intervention, whereas a lower value reflects its reliance on human assistance.

\section{Use Cases of \toolname{}}
\label{sec:3:usecase}

\toolname{} can serve as a useful tool for agentic research on understanding and annotating human-agent collaborative behavior. We particularly highlight three use cases in the current scope of this work.

\subsection{Web Automation}
\label{subsec:task1}
\toolname{} can be a standalone agent framework to automatically conduct web tasks for end users. \toolname{} is implemented as a Chrome extension where all computations other than the LLM calls are handled locally with minimal storage requirement (<50MB). Any users can easily install \toolname{} with just four clicks and use it using their personal API key. We use LiteLLM\footnote{\url{https://docs.litellm.ai/docs/}} proxy server for our backend LLM, enabling \toolname{} to support all models available via LiteLLM, including closed-source (e.g., \gpt{}) and open-weight models (e.g., \llama{}).
Depending on whether the user wants to participate in task-solving, our agent can operate in two modes: 1) \textit{Fully autonomous mode}: the agent conducts a user-issued task start-to-end; 2) \textit{CoPilot mode}: human and agent collaboratively solve a task, often useful for complex tasks where the agent is more prone to make mistakes.

\subsection{Data Collection from Websites}
\label{subsec:task2}
\toolname{} can also be used as a data annotation tool to collect task trajectories across any website accessible via the Chrome browser. Deployed as a Chrome extension, \toolname{} requires no additional setup and supports both simulated, sandboxed and self-hosted websites.

\toolname{} can track all actions conducted by LLM agents and humans. Besides, it can also collect human feedback at both (i) step-level: whether the user judges that the current step to lead to task success, and (2) task-level: whether the entire trajectory correctly solves the task.
These rich data collections can easily facilitate various studies such as user behavior studies and advanced agent learning strategies.

\begin{table*}[!htp]
\centering
\resizebox{0.9\textwidth}{!}{%
\begin{tabular}{ccccccccc}
\toprule
 \multicolumn{1}{c}{\multirow{2}{*}{\begin{tabular}[c]{@{}c@{}} \\ \\ Mode\end{tabular}}} & \multicolumn{1}{c}{\multirow{2}{*}{\begin{tabular}[c]{@{}c@{}} \\ LLM \\ Backbone\end{tabular}}} & \multirow{2}{*}{\begin{tabular}[c]{@{}c@{}}\\ End-to-End \\ Task \\ Accuracy\\ ($\uparrow$)\end{tabular}} &  & \multicolumn{5}{c}{Human-Agent Collaboration Metrics} \\ \cline{5-9} 
 & \multicolumn{1}{c}{} &  &  & \begin{tabular}[c]{@{}c@{}}Agent\\ Step \\ Count\\ ($\uparrow$)\end{tabular} & \begin{tabular}[c]{@{}c@{}}Human\\ Step\\ Count\\ ($\downarrow$)\end{tabular} & \begin{tabular}[c]{@{}c@{}}Total\\ Step\\ Count\\ ($\downarrow$)\end{tabular} & \begin{tabular}[c]{@{}c@{}}Human\\ Intervention\\ Count\\ ($\downarrow$)\end{tabular} & \begin{tabular}[c]{@{}c@{}}Agent-driven\\ Completion\\ Accuracy\\ ($\uparrow$)\end{tabular} \\ \hline
\multirow{2}{*}{\begin{tabular}[c]{@{}l@{}}Fully Autonomous\end{tabular}} & {\sc GPT-4o} & 0.48 &  & 5.48 & 0.00 & 5.48 & 0.00 & 0.48 \\
 & {\llama 8B} & 0.04 &  & 7.00 & 0.00 & 7.00 & 0.00 & 0.04 \\ \hline
\multirow{2}{*}{CoPilot} & {\sc GPT-4o} & \textbf{0.95} &  & \textbf{6.36} & \textbf{1.14} & \textbf{7.50} & \textbf{0.73} & \textbf{0.52} \\
 & {\llama 8B} & 0.81 &  & 4.77 & 4.15 & 8.92 & 1.15 & 0.05 \\ \hline
Human-only & - & 0.89 &  & 0.00 & 9.93 & 9.93 & - & - \\ \bottomrule
\end{tabular}%
}
\caption{Evaluation on WebArena subset using \toolname{}.}
\label{tab:accuracy_table}
\end{table*}

\subsection{Evaluation and Comparative Analysis of Agent Performance}  
\label{subsec:task3}  
% \faria{Rewrote this part entirely, @zora, let me know if this addresses the points you mentioned.} 
\toolname{} can be used to evaluate and compare agent performance. We support a wide range of open-weight and closed-source models served via LiteLLM. While this paper focuses on comparing \gpt and \llama, the framework can easily extend to other open and closed-source models accessible through LiteLLM.

To evaluate a particular model, the user can select a model before initiating a task. Once the task is completed, \toolname{} presents the task summary containing the results on our proposed evaluation metrics from \S\ref{sec:2.2:eval-metrics}. 
To compare different models on the same task, the user can re-do the task with different models, allowing for clear, unbiased comparisons under identical conditions. We purposefully prevent the users from changing the LLM backbone when a task is running, so that the results are not biased by the other LLMs in competition.
\section{Exemplar Findings via \toolname{}}
\label{sec:results}

To demonstrate the usage of \toolname{}, we evaluate on a subset of WebArena \citep{zhou2023webarena} benchmark, including 27 tasks categorized into easy, medium, and hard difficulty levels. We categorize the difficulty by the number of examples successfully solved by the top-performing agent \citep{wang2024agent} on WebArena, and assign them as easy, medium, hard if they have <2, 2--4, and >4 correctly solved examples among the same task template group. 
We evaluate under two settings: fully autonomous and copilot mode, using \texttt{gpt-4o-2024-08-06} and \texttt{Llama-3.1-8B-Instruct} as backbones for the LLM agent. For this study, three authors served as human agents, independently performing the tasks for both settings. The results reported represent the average performance across these evaluations.

Additionally, we included a baseline where tasks were executed solely by humans without any agent participation. \autoref{tab:accuracy_table} reports results on all metrics introduced in \S\ref{sec:2.2:eval-metrics}.

\subsection{Copilot Mode Achieves the Best Accuracy}
CoPilot mode with \textsc{GPT-4o} achieves $95\%$ task accuracy, significantly outperforming the $48\%$ accuracy under autonomous mode (relatively by $97.9\%$), and even surpassing human task-solving accuracy by $6.7\%$. This suggests potential productivity increases when solving tasks together with strong LLM-based agents.

On the other hand, copilot mode with the smaller \textsc{Llama 8B} model does not bring similar accuracy increases, but slightly degrades the task accuracy by 8\%.%, indicating the limited utility of LLM-based agents backboned by weaker LLMs.

\subsection{Copilot Mode Requires Minimal Human Intervention}

\begin{figure}
    \centering
    \includegraphics[width=0.9\linewidth]{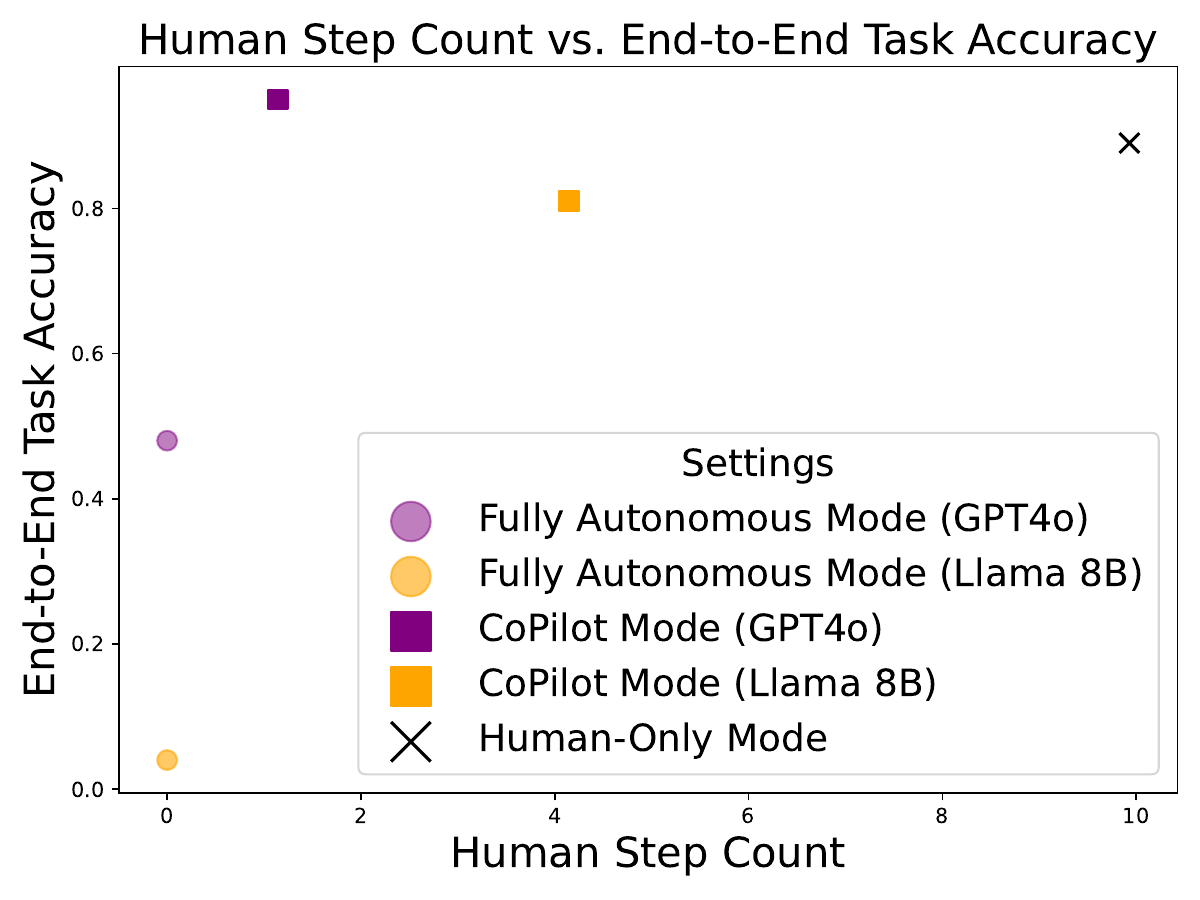}
    \caption{Correlation between Human Step Count and End-to-End Task Accuracy.
    }
    \label{fig:human_step_vs_accuracy}
\end{figure}

Despite the high task success rates, the \gpt-based agent easily achieves the highest accuracy with an average of $1.1$ human steps, taking only $15.2\%$ of the entire trajectory. Instead, the LLM agent performs the majority, more precisely $84.8\%$ of task steps.
Similarly, when shifting to the weaker \llama model, the human-llm collaboration process requires two times more human involvement, resulting in humans and LLM agents spending roughly similar amounts of effort, taking $4.77$ and $4.15$ respectively. \autoref{fig:human_step_vs_accuracy} shows the correlation between human step count and end-to-end task accuracy. 

Qualitatively, humans often choose to intervene when they observe that the LLM has gotten stuck (e.g., producing the same invalid actions multiple times) or performs an obviously wrong action (e.g., clicking `Customers' instead of `Orders' tab when searching for a particular order), especially when the webpage layout is less common or has a confusingly large number of elements.

\begin{table*}[!htp]
\centering
\vspace{-3mm}
\resizebox{0.8\textwidth}{!}{%
\begin{tabular}{lcccccc}
\toprule
 & \begin{tabular}[c]{@{}c@{}}Live \\ Website\end{tabular} & \begin{tabular}[c]{@{}c@{}}Dynamic \\ Website\end{tabular} & \begin{tabular}[c]{@{}c@{}}End-to-End  \\Human Annotation\end{tabular} & \begin{tabular}[c]{@{}c@{}}Human-Agent \\ Interaction\end{tabular} & \begin{tabular}[c]{@{}c@{}}Human-Agent  \\Co-task Execution\end{tabular} & \begin{tabular}[c]{@{}c@{}}Agent \\ Evaluation\end{tabular} \\ \midrule
WebArena & \cmark &\xmark &\xmark&\xmark&\xmark& \xmark \\
SeeAcT & \xmark & \xmark & \cmark & \xmark &\xmark& \xmark \\
BrowserGym & \cmark &\xmark&\xmark&\cmark&\xmark& \cmark \\
WebLinX &\xmark&\xmark& \cmark & \cmark &\xmark& \xmark \\
WebCanvas & \cmark & \cmark & \cmark &\xmark&\xmark& \cmark \\
WebOlympus &\cmark &\cmark & \xmark & \cmark &\xmark& \cmark \\
\textbf{\toolname{}} & \cmark & \cmark & \cmark & \cmark & \cmark & \cmark \\
\bottomrule
\end{tabular}%
}
\vspace{-1mm}
\caption{Comparison of \toolname{} with existing agent web navigation frameworks.}
\label{tab:comparison}
\end{table*}

\subsection{Agents Drive Up to Half of the Success}
In CoPilot mode, we notice that agent-drive completion accuracy was up to $52\%$ of the time with \textsc{GPT-4o} model. Note that, given the task accuracy was $0.95$, the copilot-mode agent successfully initiated half of the successes. These findings highlight that agents can follow the task objective and understand user actions to drive the task up to succeed.

\section{Related Works}

\subsection{Web Agent Plugin}
The rise of LLM agents has led to the development of open-source toolkits for web automation, available as APIs, simulated environments, and Chrome extensions. Tools like MultiOn \cite{multion} and Anthropic \cite{anthropic_2024} provide APIs for agent use but require setting up Docker images, posing barriers for non-technical users. BrowserGym \cite{drouin2024workarena}, AgentLab \cite{dechezelles2024browsergymecosystemwebagent}, WebArena \cite{zhou2023webarena} utilize a dedicated Chromium browser instance to perform tasks on specified websites. However, this approach isolates browsing sessions, restricts multi-tab navigation, and diverges from standard workflows, which limits practical usability.

Chrome extensions, as adopted by tools like WebCanvas \cite{pan2024webcanvas}, WebOlympus \cite{webolympus}, OpenWebAgent \cite{openwebagent}, and Taxy \cite{taxy_ai_2024}, present a more user-friendly alternative. They are easy to install, lightweight, and integrate seamlessly into standard browsing environments, making them accessible to end-users. While similar to \toolname{}, the extensions above lack features for fostering richer human-agent collaboration. 
% \toolname{} not only supports collaborative task execution but also incorporates mechanisms for real-time feedback collection, enabling a more interactive user experience.
\autoref{tab:comparison} further compares how \toolname{} with the existing frameworks by illustrating its novel features. The comparison is based their affordance in six categories: 1) Live, self-hosted website (\textit{e.g.},  WebArena); 2) Dynamic website that changes over time (\textit{e.g.},  shopping website); 3) Collection of end-to-end (\textit{i.e.}, human-only mode) annotation; 4) Interaction between human and agent; 5) Co-execution capability of human and agent when a task is ongoing; and 6) Interactive Evaluation of Agent performance.

\subsection{LLM Agents for Web automation}

Web automation has evolved through advancements in LLM-based agents and benchmarks. Early systems relied on HTML structures and accessibility trees \cite{deng2024mind2web, Gur2023ARW, Gur2022UnderstandingHW, Kim2023LanguageMC}. Visual-based systems such as SeeACT \cite{zheng2024gpt}, VisualWebArena \cite{koh2024visualwebarena}, WebGUM \cite{Furuta2023MultimodalWN} integrate spatial and visual understanding, enhancing agent performance in multimodal tasks. Benchmarks such as MiniWoB \cite{shi2017world} laid the foundation for evaluating these interactions, while systems like WebShop \cite{yao2022webshop}, WebArena \cite{zhou2023webarena}, WebLINX \cite{lu2024weblinx} expanded to complex multi-step tasks in e-commerce and real-world websites.

Despite these advances, existing systems focus largely on full autonomy, with limited support for human-in-the-loop collaboration. In contrast, \toolname{} bridges this gap by enabling dynamic, real-time human-agent interaction. Features like suggest-then-execute, pause, and resume facilitate adaptive task execution, make \toolname{} a robust platform for developing and evaluating agents in practical, real-world settings.

\section{Limitation and Future Work}
\label{sec:future_work}
Currently \toolname{} requires a human to act as an observer to oversee the task execution. This setup is intentional so that we can simulate task execution in \textit{live} setting. We would like to extend our work so that it does not require constant human observation. Rather, we would detect the critical steps that require human observation only. In the future, we would extend \toolname{} for a multi-LLM agent setup where we can simulate a user by a second LLM agent. Such setup would help us to approximate human decisions automatically using LLM autorater \cite{pan2024autonomous} and incorporate an active learning framework \cite{bai2024digirl}. We acknowledge a potential ordering bias in the comparative evaluation of autonomous and CoPilot modes. Moreover, due to the human-centric nature of \toolname{}'s design, the results and intervention pattern might vary based on the user. We are currently conducting a large-scale study across a diverse demographic and diverse websites to assess and mitigate the impact of such behavioral patterns.

\section*{Societal Impact}
Web agents have significant potential to promote web accessibility and enhance user efficiency. However, their deployment raises important privacy and security concerns. For instance, tracking user actions may expose sensitive information, which could be exploited for malicious purposes ({\em e.g.} data theft). Additionally, agents may inadvertently perform harmful or irreversible actions ({\em e.g.} confirming financial transactions without explicit user consent). Beyond inadvertent risks, there is also the potential for intentional misuse, where malicious actors could exploit \toolname{} for unethical purposes. We firmly discourage any such misuse of \toolname{}. To ensure safety, we have chosen not to open-source our codebase until we perform an extensive user study on \toolname{}. For now, we are only releasing the final chrome extension package. Users must provide their own API key for supported LLMs, or they can modify the extension to use a different LLM if they prefer, ensuring that their information is not shared with us. Future work can focus on addressing such safety risks and transparency, including developing robust safeguards to prevent unintended actions.

\section*{Acknowledgments}
We would like to thank Daniel Fried, Tianqi Chen, and Yonatan Bisk for their valuable feedback on the design of \toolname{}. We also acknowledge the creators of Taxy \cite{taxy_ai_2024} and BrowserGym \cite{drouin2024workarena} for open-sourcing their codebases on GitHub which served as the foundational building blocks for \toolname{}’s development.%. which provided invaluable resources for the development of \toolname{}.

\bibliography{papers}

\clearpage
\appendix

\section{Appendix}
\label{sec:appendix}

\begin{figure}
    \centering
    \includegraphics[width=0.7\linewidth]{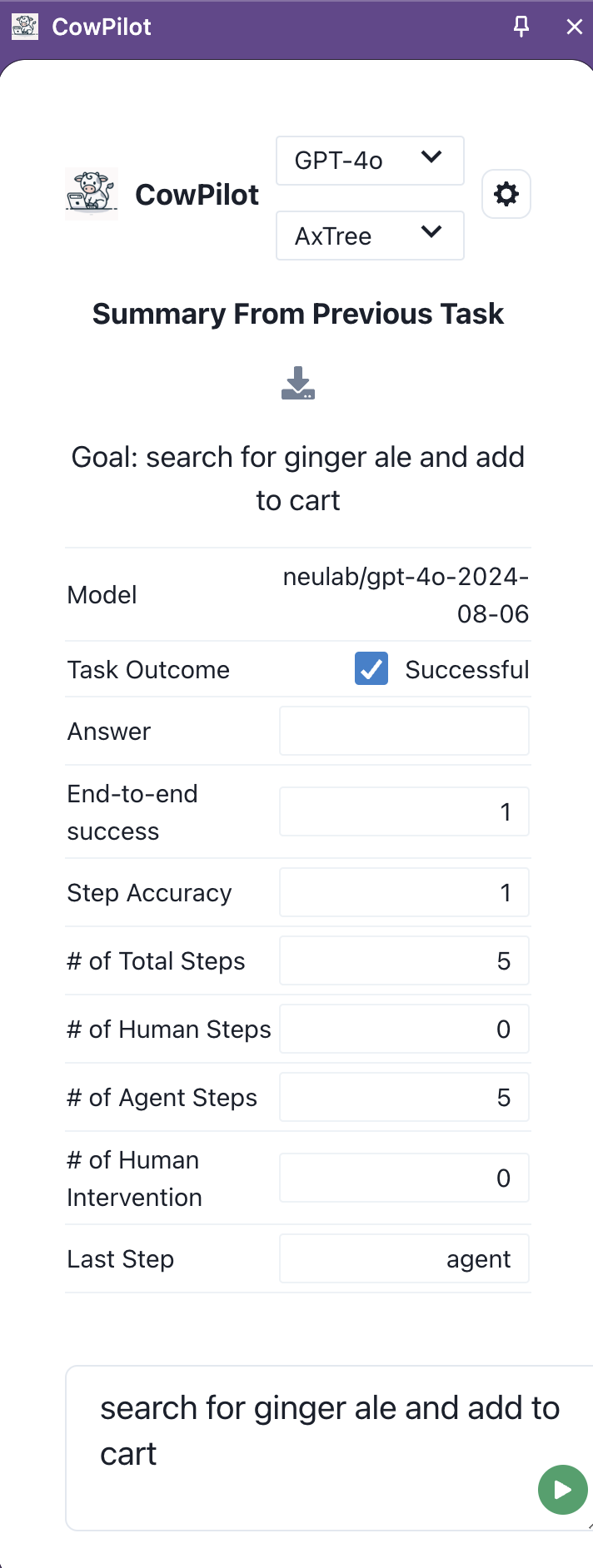}
    \caption{Screenshot of \toolname{} evaluation result page. After each task is completed, the evaluation metric values are shown as summary.}
    \label{fig:summary_table}
\end{figure}

\autoref{fig:summary_table} shows a screenshot of evaluation results by \toolname{}. After the task is completed, the summary will be shown containing the metric values covered by \autoref{sec:2.2:eval-metrics}. The scores are auto-calculated based on user intervention count and pause/resume statistics. The user can also modify each entry and save a local copy of the trajectory data by clicking on the download icon.

\subsection{Prompt for Action Transformation}
\label{subsec:trans}

\autoref{prompt-transform} shows the prompt we used for \texttt{GPT-4o-2024-08-06} to transform the raw user action into filtered actions. The prompt is provided with the event log data structure as well as the agent action space structure. These structures help the LLM to be aware of the structural representation of input raw action. The LLM replies with a list of actions together with their natural language description. The output will be used for the LLM agent action prediction when it is resumed.

\begin{figure*}
\begin{mdframed}[]
{\fontsize{7}{7}
\texttt{You will be shown a list to HTML eventlistener logdata of the following format:\\ `export interface EventLogStructure\{\\action\_type: string; // event type (click/scroll/keyup/input/KeyboardEvent/mouseover/contextmenu)\\       nodeID?: string; // if set, unique ID of the element acted on \\       elementName?: string;\\       DOM?: string;\\       elementouterHTML?: string; \\       AXTree?: string; // accessibility tree of the HTML page\\       Screenshot?: string;\\       coordinateX?: number;\\       coordinateY?: number;\\       clickType?: string;\\       position?: string;\\       URL?: string; // URL of the current page whre the events are taking place\\       scrollData?: \{\\         deltaX: number;\\         deltaY: number;\\         deltaMode: number;\\         isLine: boolean;\\         isPage: boolean;\\         isPixel: boolean;\};\\       keyData?: \{\\         key: string;\\         code: string;\\         isCtrlPressed: boolean;\\         isShiftPressed: boolean;\\         isAltPressed: boolean;\\         isMetaPressed: boolean;\\         fulltextentry: string;\};\\       urldata?: \{ // when new tab is opened, the information of the new url and tab id\\           url\_name: string;\\           tab\_id: number;\};\}`\\Your task is to clean up the raw event data and make a clean list of user actions in the following format: \colorbox{cyan}{Agent Action Space} \\   Rules:\\   1. Try to merge consecutive UserLogStructure whenever possible. For example, you can merge multiple keyup actions in the same input field as a setvalue event. For consecutive input in a textbox, always pick the final one. For example, 1) setValue(20, 'Hello') ... 10) setValue (20, 'Hello world') can be merged into a single action setValue (20, 'Hello world')\\   2. If there are repetitive user actions of the same type in the same place, feel free to discard duplicates. This might specially be true for scroll and mouseover event. For example: two consecutive scrolls in the same direction can be merged. Or, a random, disjoint scroll can be considered as a noise to be ignored.\\   3. Only reply with availableActions.name(args) format. Do not write any code.\\   4. Mouseover user log can often be noisy, only add this to the final list if it is meaningful with the rest of the action trajectory in prior and after the mouseover event. For example, a mouseover while tying into a textfield is not useful and can be discarded.\\   5. Your response must follow json format: {[}\{"thought": short summary of the action, "action": your generated action\}{]}.` \\ Input: \colorbox{green}{Raw User Actions}
}}
\end{mdframed}
\caption{Prompt for Action Transformation from Raw Event to Agent Action Space.}
\label{prompt-transform}
\end{figure*}
\vspace{0.8cm}

\end{document}